\documentclass[a4paper]{article}

\usepackage{INTERSPEECH2022}
\usepackage{multirow}

\title{ILT: Iterative LoRA Training through Focus–Feedback–Fix for Multilingual Speech Recognition}
\name{Qingliang Meng$^*$, Hao Wu$^\dagger$, Wei Liang$^*$, Wei Xu$^\dagger$, Qing Zhao$^*$}
\address{
  $^*$Megatronix (Beijing) Technology Co., Ltd\\
  $^\dagger$Shanghai Qi Zhi Institute}
\email{qingliang.meng@megatronix.co}

\begin{document}

\maketitle
\begin{abstract}
The deep integration of large language models and automatic speech recognition systems has become a promising research direction with high practical value. To address the overfitting issue commonly observed in Low-Rank Adaptation (LoRA) during the supervised fine-tuning (SFT) stage, this work proposes an innovative training paradigm Iterative LoRA Training (ILT) in combination with an Iterative Pseudo Labeling strategy, effectively enhancing the theoretical upper bound of model performance. Based on Whisper-large-v3 and Qwen2-Audio, we conduct systematic experiments using a three-stage training process: Focus Training, Feed Back Training, and Fix Training. Experimental results demonstrate the effectiveness of the proposed method. Furthermore, the MegaAIS research team applied this technique in the Interspeech 2025 Multilingual Conversational Speech Language Modeling Challenge (MLC-SLM)\footnote{https://www.nexdata.ai/competition/mlc-slm}, achieving 4\textsuperscript{th} in Track 1 (Multilingual ASR Task) and 1\textsuperscript{st} place in Track 2 (Speech Separation and Recognition Task), showcasing the practical feasibility and strong application potential of our approach.
\end{abstract}

\section{Introduction}
Multilingual automatic speech recognition (ASR) has long been a central focus in the field of speech recognition\cite{tjandra2023massively, pratap2020massively, li2021scaling, joshi2021multiple}. This area faces two core challenges: first, how to build a unified model capable of accurately recognizing speech across multiple languages and dialects; second, the scarcity of data resources for low-resource languages. To address these issues, traditional multilingual ASR models typically adopt an encoder-decoder architecture based on the Transformer framework\cite{radford2023robust, wang2025camel, peng2024owsm, meng2025dolphin}, and leverage large-scale weakly labeled data along with semi-supervised learning strategies during pretraining. This approach enables models to acquire zero-shot capabilities in low-resource languages, partially alleviating the limitations imposed by data scarcity.

In recent studies, large language models (LLMs) have achieved remarkable breakthroughs in natural language processing (NLP)\cite{touvron2023llama, bai2023qwen}. Inspired by this progress, researchers have begun exploring the application of LLMs’ strong capabilities in language understanding, generation, and representation fitting to multi-task speech processing scenarios\cite{zhang2023speechgpt}. However, prior work \cite{chu2024qwen2, xie2024mini} has primarily focused on multitask pretraining, allowing speech LLMs to support a wide range of multimodal tasks such as speech recognition, separation, and summarization. Despite this versatility, such models often lack task-specific optimization for ASR. Consequently, fine-tuning strategies tailored to ASR based on LLM pretraining have emerged as a key direction for improving multilingual ASR performance and are becoming a vital research frontier in the field.

Low-Rank Adaptation \cite{hu2022lora} (LoRA) plays a crucial role in the supervised fine-tuning (SFT) of large language models (LLMs). Existing studies primarily focus on its application to the Whisper \cite{song2024lora, xu2024towards} model, providing a promising direction for cross-lingual and low-resource dialect adaptation. HDMoLE\cite{mu2025hdmole} introduces a mixture of LoRA experts mechanism, which has been explored specifically in the domain of multi-accent speech recognition.

Pseudo Labeling Training (PL) is a form of semi-supervised learning that includes a special iterative training strategy, as discussed in studies\cite{xu2020iterative, li2023improved}. This iterative PL process involves multi-stage training in which each iteration adjusts across several dimensions, including data, model parameters, task objectives, and noise levels\cite{park2020improved, weninger2020semi}. Although such strategies are relatively complex, they offer a key advantage: the ability to guide the model toward desired performance through gradual task modulation while effectively avoiding catastrophic forgetting.

Building on prior advances in iterative LoRA-based training for text-based LLMs. For example, Li et al\cite{li2024chinchunmei}, introduced a three-stage training strategy that incorporates contrastive learning. Meanwhile, PeriodicLoRA\cite{meng2024periodiclora} investigates the iterative accumulation of low-rank update matrices, aiming to overcome performance bottlenecks and enable more effective adaptation.

Inspired by these developments, this study extends the iterative LoRA paradigm to multilingual speech recognition. Our main contributions are as follows:
\begin{itemize}
\item We propose an iterative fine-tuning paradigm based on LoRA, which leverages explicitly defined knowledge tasks to guide the optimization direction of a pretrained large language model. This paradigm overcomes the overfitting limitations commonly encountered in SFT and further improves the model’s performance ceiling.
\item We define the three-stage ILT framework, consisting of Focus, Feed Back, and Fix Training phases. By employing tailored data mixing strategies and dynamically adjusting LoRA parameters, the framework enables controlled adaptation to specific knowledge tasks throughout the training process.
\item We design strategies to identify highly relevant data and enhance the quality of pseudo texts. These strategies, applied during the Feed Back and Fix Training phases, provide high-quality supervision samples that significantly improve model performance through LoRA-based adaptation.
\end{itemize}

\section{Methodology}
\begin{figure}[htbp]
  \centering
  \includegraphics[width=\linewidth]{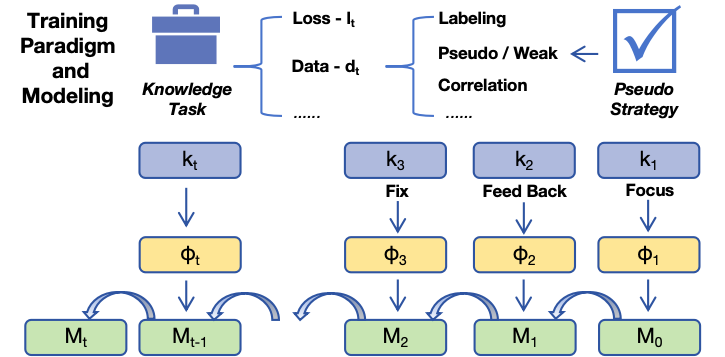}
  \caption{Training paradigm and modeling.}
  \label{fig:speech_production}
\end{figure}

In this section, we elaborate on the model training procedure and the adopted modeling methodology. A detailed overview of the workflow is presented in Figure ~\ref{fig:speech_production}.

\vspace{-10pt}
\subsection{Training Paradigm and Modeling}
Low-Rank Adaptation \cite{hu2022lora} has been widely applied in the field of natural language processing. By freezing the parameters of the original large language model and updating only the low-rank incremental weight matrices, LoRA enables efficient adaptation of the model to downstream tasks. The corresponding formulation is provided in Equation~\eqref{eq:lora_loss}. During fine-tuning, the pre-trained weights are frozen and only the low-rank matrices ${A}$ and ${B}$ are optimized, greatly reducing the number of trainable parameters.
\begin{equation}
\mathcal{L}_{\text{lora}} = - \sum_{t=1}^{T} \log P(y_t \mid y_{<t}, x; A, B)
\label{eq:lora_loss}
\end{equation}
Here, ${\mathcal{L}_{lora}}$ denotes the standard autoregressive training loss, computed over the target sequence ${y_t}$. 

Under the transfer learning framework, given a pretrained model ${M_{0}}$ and a set of downstream knowledge tasks ${k_{0}}$. we initialize the LoRA parameters and optimize them on ${k_{0}}$ guided by a loss function. After fine-tuning with LoRA, the adapted model weight ${M_{1}}$ is obtained by merging the base model weight ${M_{0}}$ with the learned low-rank update ${BA}$. Optionally, a scaling factor ${\frac{\alpha}{r}}$ can be applied to control the update magnitude.
\begin{equation}
M_1 = \phi_1 M_0 = M_0 + \Delta M = M_0 + \frac{\alpha}{r} BA
\end{equation}
where are the learnable LoRA parameters ${A \in \mathbb{R}^{r \times d_{\text{in}}}}$, ${B \in \mathbb{R}^{d_{\text{out}} \times r}}$, ${r}$ is the low-rank dimension, and ${\alpha}$ is a scaling factor, $\phi_1$ denote the coefficients capturing the effect of historical values on the present.

This study draws inspiration from the theory of Deliberate Practice \cite{anders2008deliberate} in behavioral science, which emphasizes structured and repetitive training. For example, pianists decompose complex compositions into smaller segments for targeted practice, and athletes enhance their skills through multiple rounds of focused physical and technical training. Based on this concept, we propose the Iterative LoRA Training (ILT) paradigm. The ILT approach assumes that the optimization of LoRA parameters follows a discrete time iterative process and gradually converges toward a global optimum through multiple training stages. The modeling process is described in Equation~\eqref{eq:ilt_loss}, where a recursive optimization mechanism enables continuous improvement in model performance.

\begin{equation}
\phi_t = \arg\min_{\phi_t} \ \mathcal{L}_{k_t}, \quad 
M_t = \phi_t M_{t-1} + \cdots + \phi_{t-n+1} M_{t-n}
\label{eq:ilt_loss}
\end{equation}

Here, ${t \in \{1, 2, \dots, n\}}$ denotes the iteration index, At each iteration $t$, ${k_{t}}$ refers to the selected knowledge task, ${L_{k_{t}}}$ represents the optimization objective for this task, and $\phi_t$ denotes the updated parameters obtained through LoRA-based fine-tuning.

Specifically, in this MLC-SLM challenge, we conducted three rounds of ILT, which we refer to as Focus Training, Feed Back Training, and Fix Training. Detailed descriptions of these stages will be provided in the Experiment~\ref{sec:tsits} section.

\subsection{Knowledge Task Design}
The knowledge task set ${K}$ consists of a series of task units ${k_{t}}$, where each task ${k_{t}}$ is composed of a loss objective ${l_{t}}$ and a dataset ${d_{t}}$. Formally, the set is defined as follows.

\begin{equation}
K = \{k_t = [l_t, d_t]\}_{t=1}^N
\end{equation}
For each ${k_{t}}$, the model is trained on dataset ${d_{t}}$ with the corresponding loss function ${l_{t}}$ as the optimization target. Different training stages may be assigned distinct optimization objectives and corresponding datasets, depending on the learning requirements at each stage, in order to achieve optimal performance.

To further enhance the model’s performance in multilingual automatic speech recognition tasks, we designed a data augmentation and selection pipeline tailored for the construction of knowledge tasks. All tasks share a unified optimization objective based on a generative loss~\eqref{eq:lora_loss}.

This augmentation and selection pipeline is applied only during the Feed Back Training and Fix Training stages. It integrates multiple strategies, including speech synthesis models, acoustic feature matching, text token distribution analysis, and model-driven error analysis based on performance feedback. Detailed implementation procedures can be found in the Experiment~\ref{sec:ekt} section.

\subsection{Improving the Reliability of Pseudo Labeling}
\label{sec:pl}
To improve the reliability of pseudo references introduced during the Feed Back and Fix Training stages, we adopt a hard voting strategy based on ensemble learning. For each audio sample, ${N}$ models, either from different training phases or from various optimization steps within the same phase, generate candidate hypotheses. The final pseudo reference is determined by majority voting\cite{burka2022voting}, which helps ensure higher transcription reliability.
\begin{equation}
{ref}_{pseudo} = \text{Vote}\left( \{hyp_{n} = M_n(x) \}_{n=1}^{N} \right)
\end{equation}
where ${Vote()}$ denotes majority voting across predicted sequences.

The pseudo labels generated through this mechanism are continuously incorporated into the knowledge task set ${K}$. This ensures that the quality of pseudo labels remains aligned with the current stage of model training and contributes to performance improvement across iterations.

\section{Experiment}
\label{sec:exp}
\subsection{Three-Stage Iterative Training Strategy}
\label{sec:tsits}
Considering the time constraints of the competition, we employed only three iterations of Iterative LoRA Training~\eqref{eq:ilt_loss}. Specifically, the training process consisted of three consecutive stages: Focus Training, Feed Back Training, and Fix Training.

\begin{table*}[htbp]
\centering
\caption{Comparison results of different Systems(SD: Speaker Diarization. ILT:  Iterative LoRA Training)}
\vspace{-5pt}
\label{tab:comSystems}
\begin{tabular}{lcccccccc}
\toprule
\multirow{2}{*}{\textbf{Language}} & 
\multicolumn{4}{c}{\textbf{Baseline Models (WER\%)}} & 
\multicolumn{2}{c}{\textbf{ILT Models (WER\%)}} &
\multicolumn{2}{c}{\textbf{SD Models (tcpWER\%)}} \\
\cmidrule(lr){2-5} \cmidrule(lr){6-7} \cmidrule{8-9}
& \textbf{Whisper} & \textbf{Qwen} & \textbf{Llama} & \textbf{Qwen2-Audio} & \textbf{MegaWhisper} & \textbf{MegaAudio} & \textbf{Baseline} & \textbf{MegaAudio}\\
\midrule
English-American     & 14.14 & 14.04 & 17.45 & 12.58 & 9.98  & 10.29 & 53.73 & 24.23 \\
English-Australian   & 11.72 & 11.60 & 13.77 & 13.77 & 6.01  & 6.19 & 52.63 & 13.38 \\
English-British      & 10.08 & 11.37 & 12.09 & 12.32 & 6.16  & 6.61 & 71.92 & 21.90 \\
English-Filipino     & 9.20  & 8.15  & 8.43  & 14.94 & 6.18  & 6.95 & 50.37 & 14.21 \\
English-Indian       & 13.96 & 17.73 & 16.22 & 21.66 & 9.45  & 11.28 & 70.72 & 10.62 \\
French               & 26.72 & 25.33 & 26.84 & 70.61 & 13.80 & 14.73 & 96.04 & 28.41 \\
German               & 20.53 & 36.64 & 34.06 & 91.52 & 17.04 & 17.98 & 86.74 & 27.62 \\
Italian              & 17.94 & 24.22 & 24.91 & 70.95 & 11.14 & 12.27 & 83.31 & 15.85 \\
Japanese             & 21.64 & 34.88 & 34.06 & 21.02 & 14.25 & 14.57 & 71.30 & 27.23 \\
Korean               & 13.80 & 20.60 & 22.31 & 37.85 & 8.58  & 9.27 & 59.55 & 24.79 \\
Portuguese           & 20.82 & 36.09 & 33.82 & 115.86 & 16.73 & 18.06 & 118.84 & 28.68 \\
Russian              & 7.36  & 7.51  & 7.97  & 62.86 & 2.82  & 3.49 & 69.21 & 14.81 \\
Spanish              & 12.24 & 15.00 & 17.03 & 91.85 & 8.21  & 8.32 & 75.61 & 16.88 \\
Thai                 & 14.49 & 23.10 & 19.98 & 101.03 & 8.21  & 9.00 & 83.56 & 19.89 \\
Vietnamese           & 23.02 & 18.22 & 19.66 & 103.28 & 11.29 & 11.76 & 82.80 & 23.89 \\
\midrule
Avg                  & 15.36 & 19.82 & 19.74 & 51.90 & 9.20  & 9.89 & 76.12 & 21.32 \\
\bottomrule
\end{tabular}
\vspace{-8pt}
\end{table*}

\vspace{-8pt}
\subsubsection{Focus Training}
Speech large language models are typically pretrained using large scale audio datasets. For example, Whisper \cite{radford2023robust} relies extensively on pseudo-labeled and semi-supervised audio data during its pretraining. In contrast, Qwen2-Audio \cite{chu2024qwen2} is designed to support a broad range of audio tasks such as speech summarization and classification, rather than focusing exclusively on automatic speech recognition.

In this study, Focus Training serves as the first stage, aiming to rapidly adapt a pretrained model to the target task and data domain. To avoid catastrophic forgetting of prior knowledge during fast adaptation, a small scale parameter tuning strategy is applied. Specifically, the LoRA parameters are configured with \(lora\_alpha\) set to 32 and \(lora\_r\) set to 16. The target modules selected for adaptation include \(q\_proj\), \(k\_proj\), and \(v\_proj\).

The model is trained for six epochs on the official MLC-SLM dataset and evaluated on the development set. The best performing LoRA parameters are merged with the base model to obtain an adapted version, referred to as ${M_1}$.
\vspace{-5pt}
\subsubsection{Feed Back Training}
After Focus Training, the merged model ${M_1}$ presents two main problems. First, the model quickly converges to specific domains, which increases the risk of overfitting during continued training. Second, although LLM based speech models are expected to have zero-shot generalization capability for unseen languages, the actual performance remains unsatisfactory. For example, Qwen2-Audio has not been pretrained on Thai and Vietnamese, leading to poor recognition results in these languages due to missing language specific knowledge.

To address these challenges, two improvements are implemented. The first is data expansion. The training set is enlarged to 40,000 hours by randomly selecting additional samples, and supplementary Thai and Vietnamese data are incorporated. The second is parameter scope extension. To promote learning of new knowledge, a wider range of parameters is made trainable. The target modules now include \(q\_proj\), \(k\_proj\), \(v\_proj\), \(o\_proj\), \(down\_proj\), \(gate\_proj\), \(up\_proj\), \(out\_proj\), \(fc1\), and \(fc2\). The LoRA parameters are set to \(lora\_alpha\) = 2048 and \(lora\_r\) = 512.

The model is trained for ten epochs and evaluated on the development set. The optimal LoRA parameters are merged with ${M_2}$ to produce the updated model, referred to as ${M_2}$.

\subsubsection{Fix Training}
After the Feed Back Training stage, model ${M_2}$ has undergone refinement with broader multilingual data. In the final training phase, high-quality data are selected following the strategy described in Section~\ref{sec:ekt}, resulting in 2,791 hours of relevant audio. Additionally, unlabeled audio samples are continuously pseudo-labeled using the method introduced in Section~\ref{sec:pl}.

During this phase, LoRA tuning is applied again with \(lora\_alpha\) = 64 and \(lora\_r\) = 32. The target modules remain the same as those used in the Feed Back Training stage. The model is trained for six epochs and evaluated on the development set. The optimal LoRA parameters are then merged into model ${M_2}$, completing the training process.

\subsection{Enhancing Knowledge Tasks}
\label{sec:ekt}
\subsubsection{Text Augmentation Strategy}
\label{sec:tas}
This study systematically analyzes the official training set using the Whisper-large-v3 tokenizer to compute token frequency distributions. Tokens occurring fewer than 20 times are categorized as low-frequency, while those appearing more than 200 times are considered high-frequency.

For low-frequency tokens, third-party text data are selected when such tokens account for more than 50 percent of a sentence. This approach enhances the model's capacity to learn rare vocabulary, named entities, and uncommon grammatical patterns, addressing representation imbalance caused by uneven data distribution.

For high-frequency tokens, sentences are first transcribed using Whisper-large-v3. Samples containing these tokens with a word error rate (WER) above 5 percent are flagged as difficult cases. Introducing these samples helps the model better handle frequent contexts, improving its robustness and generalization through targeted error correction.

\subsubsection{Audio Augmentation Strategy}
In terms of audio augmentation, this study designs different strategies for different languages.

For English dialects, F5-TTS\cite{chen2024f5} and Spark-TTS\cite{wang2025spark} are employed to synthesize additional training samples. Original recordings from the MLC-SLM dataset serve as prompts and are paired with low-frequency text data (as described in Section~\ref{sec:tas}). The generated audio retains phonetic consistency with the originals, enhancing data diversity in dialectal English scenarios.

For non-English languages, especially those that are low-resource or less commonly spoken, we propose a similarity-based audio selection method. Audio embeddings are extracted using Wavlm\cite{chen2022wavlm} from both the official and third-party datasets. Third-party segments with a Speaker Encoder Cosine Similarity (SECS) \cite{casanova2104sc} above 0.85 are selected. This approach does not rely on transcriptions and efficiently identifies high-quality supplementary data for rare languages, thereby supporting multilingual model development.

\subsubsection{Experimental Details and General Configuration}
We trained our models using eight A800 GPUs with a micro batch size of 16, a learning rate of 1e-5, and gradient accumulation steps set to 4. The training was conducted using bfloat16 precision. Additionally, to reduce transcription errors caused by language confusion, learnable language identifier embeddings are introduced at both the encoder and decoder output layers during ASR training.

For ASR decoding, we adopt the Beam Search algorithm, using a beam size of 6 for most languages\cite{meng2025mtlm}. Exceptions are made for Thai, Japanese, and Vietnamese, where a beam size of 1 demonstrates better empirical performance. For the speech diarization task, we rely solely on the official baseline techniques. To support ASR inference, audio segments longer than 30 seconds are evenly split into shorter chunks. 

To extend the training data, we incorporated resources from eight publicly available datasets: Multilingual LibriSpeech\cite{pratap2020mls}, Multilingual TEDx\cite{salesky2021multilingual}, Common Voice\cite{ardila2019common}, GigaSpeech 2\cite{yang2024gigaspeech}, Maxseats\cite{maxseatsmeetingvalid}, YODAS2\cite{li2023yodas}, ReazonSpeech\cite{fujimoto2016reazonspeech}, and SBCSAE\cite{dubois_2005}. These datasets span a variety of languages and application scenarios, offering strong linguistic diversity and high audio quality. This foundation supports the development of speech recognition models that are both robust and generalizable.
\vspace{-10pt}
\section{Results}
This section evaluates the effectiveness of the proposed approach and conducts ablation studies on each training stage. All experiments are carried out on the development set of the MLC-SLM benchmark \footnote{https://github.com/mubingshen/MLC-SLM-Baseline/tree/main}. To distinguish the ILT-adapted models from their original versions, we refer to the Whisper-large-v3 model after ILT training as MegaWhisper, and the Qwen2-Audio counterpart as MegaAudio in the remainder of this paper.

\subsection{Experimental Results}
Table~\ref{tab:comSystems} demonstrates that the proposed ILT fine tuning method leads to consistent performance improvements across all evaluated languages. Qwen2-Audio, which exhibits limitations due to incomplete multilingual coverage and information loss during its original pretraining, shows noticeable gains after applying ILT. In contrast, Whisper-large-v3, which was pretrained with a strong focus on automatic speech recognition, achieves even better results after ILT adaptation and consistently outperforms the Qwen2-Audio.

\subsection{Ablation Study}
The ablation study yields the following key findings. First, applying only Focus Training or Fix Training enables the model to quickly adapt to the data distribution, resulting in an initial performance improvement. Notably, Fix Training, which incorporates highly relevant data as described in Section~\ref{sec:ekt}, contributes an additional gain of 0.31 percent. Second, for Whisper-large-v3, which already contains sufficient multilingual pretraining information, introducing Feed Back Training alone may lead to a mismatch between the training and evaluation domains and cause performance degradation. In contrast, Qwen2-Audio benefits from Feed Back Training, as it helps compensate for the absence of pretraining on certain languages and thus improves recognition accuracy. Third, combining Focus Training and Fix Training without introducing additional data during the Feed Back stage results in performance decline, indicating that consecutive training on highly similar data can lead to overfitting. Finally, the proposed three-stage training strategy achieves the best overall performance in the ablation study, demonstrating the effectiveness of the iterative training design.
\vspace{-10pt}
\begin{table}[htbp]
  \centering
  \caption{Ablation Study Results}
  \vspace{-5pt}
  \label{tab:iltAblation}
  \begin{tabular}{lcc}
    \toprule
    \multirow{2}{*}{\textbf{Training Method}} & 
    \multicolumn{2}{c}{\textbf{ILT Models (WER\%)}} \\
    \cmidrule(r){2-3} 
    & \textbf{MegaWhisper} & \textbf{MegaAudio} \\
    \midrule
    Only Focus          & 10.38 & 19.92 \\
    Only Feed Back      & 11.59 & 14.53 \\
    Only Fix            & 10.07 & 18.99 \\
    Focus + Feed Back   & 9.89  & 12.15 \\
    Feed Back + Fix     & 9.83  & 10.33 \\
    Focus + Fix         & 10.62 & 19.86 \\
    \textbf{Focus + Feed Back + Fix} & \textbf{9.20} & \textbf{9.89} \\
    \bottomrule
  \end{tabular}
  \vspace{-10pt}
\end{table}

\vspace{-5pt}
\section{Conclusions}
This study proposes an iterative fine-tuning framework based on Low Rank Adaptation, aiming to enhance model performance through a knowledge task oriented adjustment mechanism. In the MLC-SLM challenge, we design a three-stage training paradigm consisting of Focus, Feed Back, and Fix Training. Each stage is carefully aligned with a specific knowledge task through optimized data ratio strategies and dynamic parameter adaptation. Furthermore, we introduce text and audio augmentation strategies to support the discovery of relevant third party data. A voting mechanism is also employed to improve the reliability of pseudo labels. The experimental section presents detailed implementation procedures, quantitative results, and ablation analyses.
\vspace{-5pt}
\bibliographystyle{IEEEtran}
\bibliography{LaTeX/template}

\begin{thebibliography}{10}
\providecommand{\url}[1]{#1}
\csname url@samestyle\endcsname
\providecommand{\newblock}{\relax}
\providecommand{\bibinfo}[2]{#2}
\providecommand{\BIBentrySTDinterwordspacing}{\spaceskip=0pt\relax}
\providecommand{\BIBentryALTinterwordstretchfactor}{4}
\providecommand{\BIBentryALTinterwordspacing}{\spaceskip=\fontdimen2\font plus
\BIBentryALTinterwordstretchfactor\fontdimen3\font minus \fontdimen4\font\relax}
\providecommand{\BIBforeignlanguage}[2]{{%
\expandafter\ifx\csname l@#1\endcsname\relax
\typeout{** WARNING: IEEEtran.bst: No hyphenation pattern has been}%
\typeout{** loaded for the language `#1'. Using the pattern for}%
\typeout{** the default language instead.}%
\else
\language=\csname l@#1\endcsname
\fi
#2}}
\providecommand{\BIBdecl}{\relax}
\BIBdecl

\bibitem{tjandra2023massively}
A.~Tjandra, N.~Singhal, D.~Zhang, O.~Kalinli, A.~Mohamed, D.~Le, and M.~L. Seltzer, ``Massively multilingual asr on 70 languages: Tokenization, architecture, and generalization capabilities,'' in \emph{ICASSP 2023-2023 IEEE International Conference on Acoustics, Speech and Signal Processing (ICASSP)}.\hskip 1em plus 0.5em minus 0.4em\relax IEEE, 2023, pp. 1--5.

\bibitem{pratap2020massively}
V.~Pratap, A.~Sriram, P.~Tomasello, A.~Hannun, V.~Liptchinsky, G.~Synnaeve, and R.~Collobert, ``Massively multilingual asr: 50 languages, 1 model, 1 billion parameters,'' \emph{arXiv preprint arXiv:2007.03001}, 2020.

\bibitem{li2021scaling}
B.~Li, R.~Pang, T.~N. Sainath, A.~Gulati, Y.~Zhang, J.~Qin, P.~Haghani, W.~R. Huang, M.~Ma, and J.~Bai, ``Scaling end-to-end models for large-scale multilingual asr,'' in \emph{2021 IEEE Automatic Speech Recognition and Understanding Workshop (ASRU)}.\hskip 1em plus 0.5em minus 0.4em\relax IEEE, 2021, pp. 1011--1018.

\bibitem{joshi2021multiple}
V.~Joshi, A.~Das, E.~Sun, R.~R. Mehta, J.~Li, and Y.~Gong, ``Multiple softmax architecture for streaming multilingual end-to-end asr systems.'' in \emph{Interspeech}, 2021, pp. 1767--1771.

\bibitem{radford2023robust}
A.~Radford, J.~W. Kim, T.~Xu, G.~Brockman, C.~McLeavey, and I.~Sutskever, ``Robust speech recognition via large-scale weak supervision,'' in \emph{International conference on machine learning}.\hskip 1em plus 0.5em minus 0.4em\relax PMLR, 2023, pp. 28\,492--28\,518.

\bibitem{wang2025camel}
H.~Wang, X.~Wan, N.~Zheng, K.~Liu, H.~Zhou, G.~Li, and L.~Xie, ``Camel: Cross-attention enhanced mixture-of-experts and language bias for code-switching speech recognition,'' in \emph{ICASSP 2025-2025 IEEE International Conference on Acoustics, Speech and Signal Processing (ICASSP)}.\hskip 1em plus 0.5em minus 0.4em\relax IEEE, 2025, pp. 1--5.

\bibitem{peng2024owsm}
Y.~Peng, J.~Tian, W.~Chen, S.~Arora, B.~Yan, Y.~Sudo, M.~Shakeel, K.~Choi, J.~Shi, X.~Chang \emph{et~al.}, ``Owsm v3. 1: Better and faster open whisper-style speech models based on e-branchformer,'' \emph{arXiv preprint arXiv:2401.16658}, 2024.

\bibitem{meng2025dolphin}
Y.~Meng, J.~Li, G.~Lin, Y.~Pu, G.~Wang, H.~Du, Z.~Shao, Y.~Huang, K.~Li, and W.-Q. Zhang, ``Dolphin: A large-scale automatic speech recognition model for eastern languages,'' \emph{arXiv preprint arXiv:2503.20212}, 2025.

\bibitem{touvron2023llama}
H.~Touvron, T.~Lavril, G.~Izacard, X.~Martinet, M.-A. Lachaux, T.~Lacroix, B.~Rozi{\`e}re, N.~Goyal, E.~Hambro, F.~Azhar \emph{et~al.}, ``Llama: Open and efficient foundation language models,'' \emph{arXiv preprint arXiv:2302.13971}, 2023.

\bibitem{bai2023qwen}
J.~Bai, S.~Bai, Y.~Chu, Z.~Cui, K.~Dang, X.~Deng, Y.~Fan, W.~Ge, Y.~Han, F.~Huang \emph{et~al.}, ``Qwen technical report,'' \emph{arXiv preprint arXiv:2309.16609}, 2023.

\bibitem{zhang2023speechgpt}
D.~Zhang, S.~Li, X.~Zhang, J.~Zhan, P.~Wang, Y.~Zhou, and X.~Qiu, ``Speechgpt: Empowering large language models with intrinsic cross-modal conversational abilities,'' \emph{arXiv preprint arXiv:2305.11000}, 2023.

\bibitem{chu2024qwen2}
Y.~Chu, J.~Xu, Q.~Yang, H.~Wei, X.~Wei, Z.~Guo, Y.~Leng, Y.~Lv, J.~He, J.~Lin \emph{et~al.}, ``Qwen2-audio technical report,'' \emph{arXiv preprint arXiv:2407.10759}, 2024.

\bibitem{xie2024mini}
Z.~Xie and C.~Wu, ``Mini-omni: Language models can hear, talk while thinking in streaming,'' \emph{arXiv preprint arXiv:2408.16725}, 2024.

\bibitem{hu2022lora}
E.~J. Hu, Y.~Shen, P.~Wallis, Z.~Allen-Zhu, Y.~Li, S.~Wang, L.~Wang, W.~Chen \emph{et~al.}, ``Lora: Low-rank adaptation of large language models.'' \emph{ICLR}, vol.~1, no.~2, p.~3, 2022.

\bibitem{song2024lora}
Z.~Song, J.~Zhuo, Y.~Yang, Z.~Ma, S.~Zhang, and X.~Chen, ``Lora-whisper: Parameter-efficient and extensible multilingual asr,'' \emph{arXiv preprint arXiv:2406.06619}, 2024.

\bibitem{xu2024towards}
T.~Xu, K.~Huang, P.~Guo, Y.~Zhou, L.~Huang, H.~Xue, and L.~Xie, ``Towards rehearsal-free multilingual asr: A lora-based case study on whisper,'' \emph{arXiv preprint arXiv:2408.10680}, 2024.

\bibitem{mu2025hdmole}
B.~Mu, K.~Wei, Q.~Shao, Y.~Xu, and L.~Xie, ``Hdmole: Mixture of lora experts with hierarchical routing and dynamic thresholds for fine-tuning llm-based asr models,'' in \emph{ICASSP 2025-2025 IEEE International Conference on Acoustics, Speech and Signal Processing (ICASSP)}.\hskip 1em plus 0.5em minus 0.4em\relax IEEE, 2025, pp. 1--5.

\bibitem{xu2020iterative}
Q.~Xu, T.~Likhomanenko, J.~Kahn, A.~Hannun, G.~Synnaeve, and R.~Collobert, ``Iterative pseudo-labeling for speech recognition,'' \emph{arXiv preprint arXiv:2005.09267}, 2020.

\bibitem{li2023improved}
T.~Li, Q.~Meng, and Y.~Sun, ``Improved noisy iterative pseudo-labeling for semi-supervised speech recognition,'' in \emph{2022 IEEE Spoken Language Technology Workshop (SLT)}.\hskip 1em plus 0.5em minus 0.4em\relax IEEE, 2023, pp. 167--173.

\bibitem{park2020improved}
D.~S. Park, Y.~Zhang, Y.~Jia, W.~Han, C.-C. Chiu, B.~Li, Y.~Wu, and Q.~V. Le, ``Improved noisy student training for automatic speech recognition,'' \emph{arXiv preprint arXiv:2005.09629}, 2020.

\bibitem{weninger2020semi}
F.~Weninger, F.~Mana, R.~Gemello, J.~Andr{\'e}s-Ferrer, and P.~Zhan, ``Semi-supervised learning with data augmentation for end-to-end asr,'' \emph{arXiv preprint arXiv:2007.13876}, 2020.

\bibitem{li2024chinchunmei}
T.~Li, N.~Rusnachenko, and H.~Liang, ``Chinchunmei at wassa 2024 empathy and personality shared task: Boosting llm’s prediction with role-play augmentation and contrastive reasoning calibration,'' in \emph{Proceedings of the 14th Workshop on Computational Approaches to Subjectivity, Sentiment, \& Social Media Analysis}, 2024, pp. 385--392.

\bibitem{meng2024periodiclora}
X.~Meng, D.~Dai, W.~Luo, Z.~Yang, S.~Wu, X.~Wang, P.~Wang, Q.~Dong, L.~Chen, and Z.~Sui, ``Periodiclora: Breaking the low-rank bottleneck in lora optimization,'' \emph{arXiv preprint arXiv:2402.16141}, 2024.

\bibitem{anders2008deliberate}
K.~Anders~Ericsson, ``Deliberate practice and acquisition of expert performance: a general overview,'' \emph{Academic emergency medicine}, vol.~15, no.~11, pp. 988--994, 2008.

\bibitem{burka2022voting}
D.~Burka, C.~Puppe, L.~Szepesv{\'a}ry, and A.~Tasn{\'a}di, ``Voting: A machine learning approach,'' \emph{European Journal of Operational Research}, vol. 299, no.~3, pp. 1003--1017, 2022.

\bibitem{chen2024f5}
Y.~Chen, Z.~Niu, Z.~Ma, K.~Deng, C.~Wang, J.~Zhao, K.~Yu, and X.~Chen, ``F5-tts: A fairytaler that fakes fluent and faithful speech with flow matching,'' \emph{arXiv preprint arXiv:2410.06885}, 2024.

\bibitem{wang2025spark}
X.~Wang, M.~Jiang, Z.~Ma, Z.~Zhang, S.~Liu, L.~Li, Z.~Liang, Q.~Zheng, R.~Wang, X.~Feng \emph{et~al.}, ``Spark-tts: An efficient llm-based text-to-speech model with single-stream decoupled speech tokens,'' \emph{arXiv preprint arXiv:2503.01710}, 2025.

\bibitem{chen2022wavlm}
S.~Chen, C.~Wang, Z.~Chen, Y.~Wu, S.~Liu, Z.~Chen, J.~Li, N.~Kanda, T.~Yoshioka, X.~Xiao \emph{et~al.}, ``Wavlm: Large-scale self-supervised pre-training for full stack speech processing,'' \emph{IEEE Journal of Selected Topics in Signal Processing}, vol.~16, no.~6, pp. 1505--1518, 2022.

\bibitem{casanova2104sc}
E.~Casanova, C.~Shulby, E.~G{\"o}lge, N.~M{\"u}ller, F.~de~Oliveira, A.~Junior, A.~da~Silva~Soares, S.~Aluisio, and M.~Ponti, ``Sc-glowtts: An efficient zero-shot multi-speaker text-to-speech model. arxiv 2021,'' \emph{arXiv preprint arXiv:2104.05557}, 2021.

\bibitem{meng2025mtlm}
Q.~Meng, P.~Ren, T.~Li, and C.~Dai, ``Mtlm: an innovative language model training paradigm for asr,'' \emph{arXiv preprint arXiv:2502.10058}, 2025.

\bibitem{pratap2020mls}
V.~Pratap, Q.~Xu, A.~Sriram, G.~Synnaeve, and R.~Collobert, ``Mls: A large-scale multilingual dataset for speech research,'' \emph{arXiv preprint arXiv:2012.03411}, 2020.

\bibitem{salesky2021multilingual}
E.~Salesky, M.~Wiesner, J.~Bremerman, R.~Cattoni, M.~Negri, M.~Turchi, D.~W. Oard, and M.~Post, ``The multilingual tedx corpus for speech recognition and translation,'' \emph{arXiv preprint arXiv:2102.01757}, 2021.

\bibitem{ardila2019common}
R.~Ardila, M.~Branson, K.~Davis, M.~Henretty, M.~Kohler, J.~Meyer, R.~Morais, L.~Saunders, F.~M. Tyers, and G.~Weber, ``Common voice: A massively-multilingual speech corpus,'' \emph{arXiv preprint arXiv:1912.06670}, 2019.

\bibitem{yang2024gigaspeech}
Y.~Yang, Z.~Song, J.~Zhuo, M.~Cui, J.~Li, B.~Yang, Y.~Du, Z.~Ma, X.~Liu, Z.~Wang \emph{et~al.}, ``Gigaspeech 2: An evolving, large-scale and multi-domain asr corpus for low-resource languages with automated crawling, transcription and refinement,'' \emph{arXiv preprint arXiv:2406.11546}, 2024.

\bibitem{maxseatsmeetingvalid}
Maxseats, ``Maxseats meeting valid dataset,'' \url{https://huggingface.co/datasets/maxseats/meeting\_valid}.

\bibitem{li2023yodas}
X.~Li, S.~Takamichi, T.~Saeki, W.~Chen, S.~Shiota, and S.~Watanabe, ``Yodas: Youtube-oriented dataset for audio and speech,'' in \emph{2023 IEEE Automatic Speech Recognition and Understanding Workshop (ASRU)}.\hskip 1em plus 0.5em minus 0.4em\relax IEEE, 2023, pp. 1--8.

\bibitem{fujimoto2016reazonspeech}
Y.~Y. D. M.~S. Fujimoto, ``Reazonspeech: A free and massive corpus for japanese asr,'' \emph{null}, 2016.

\bibitem{dubois_2005}
J.~W. Du~Bois, W.~L. Chafe, C.~Meyer, S.~A. Thompson, R.~Englebretson, and N.~Martey, ``{S}anta {B}arbara corpus of spoken {A}merican {E}nglish, {P}arts 1--4,'' Linguistic Data Consortium, Philadelphia, 2000--2005.

\end{thebibliography}

\end{document}